\documentclass[11pt]{article}
\usepackage[latin9]{inputenc}
\usepackage{float}
\usepackage{varwidth}
\usepackage{amsmath}
\usepackage{amsthm}
\usepackage{amssymb}
\usepackage[a4paper]{geometry}
\makeatletter

\providecommand{\tabularnewline}{\\}
\newenvironment{cellvarwidth}[1][t]
    {\begin{varwidth}[#1]{\linewidth}}
    {\@finalstrut\@arstrutbox\end{varwidth}}

\newenvironment{lyxlist}[1]
	{\begin{list}{}
		{\settowidth{\labelwidth}{#1}
		 \setlength{\leftmargin}{\labelwidth}
		 \addtolength{\leftmargin}{\labelsep}
		 }}
	{\end{list}}
\theoremstyle{plain}
\newtheorem{thm}{\protect\theoremname}
\theoremstyle{plain}
\newtheorem{lem}[thm]{\protect\lemmaname}

\usepackage{paralist}

\newtheorem{theorem}{Theorem}

\newtheorem{remark}{Remark}

\input{tcilatex}
\AtBeginDocument{} 

\makeatother

\providecommand{\lemmaname}{Lemma}
\providecommand{\theoremname}{Theorem}

\begin{document}
\title{Supervaluationism about vagueness and indeterminacy:\\
A judgment-aggregation perspective\thanks{Earlier versions of this paper were presented on several occasions,
including the VII Navarra Workshop on Vagueness, Universidad de Navarra,
Pamplona, December 2013, a conference on \textquotedblleft Kinds of
Indeterminacy\textquotedblright , University of Geneva, November 2016,
and the SIFA conference, University of Eastern Piedmont, Novara, September
2018. We thank the participants and especially Pablo Cobreros and
Ellie Ripley for comments. This work was supported by a Leverhulme
Major Research Fellowship of the Leverhulme Trust and the PLEXUS project
(Grant Agreement No. 101086295), funded by the EU through the Marie
Sk\l odowska-Curie Actions under the Horizon Europe Research and Innovation
Programme. PE also acknowledges the project ``Paradoxes as Philosophical
Tools'' (PID2023-150066NB-I00) and an honorary fellowship from the
University of Melbourne (Department of Electrical Engineering).}}
\author{Christian List \\
{\small LMU Munich} \and Paul Egré \\
{\small IRL Crossing, CNRS \& ENS }\and Franz Dietrich\\
{\small PSE\ \&\ CNRS}}
\date{{\small 20 November 2026}}
\maketitle
\begin{abstract}
\noindent Supervaluationism about vagueness and indeterminacy is based
on the idea that some sentences, or predicates within them, have multiple
competing interpretations or specifications. A sentence is deemed
determinately true if and only if it is true according to all specifications;
and determinately false if and only if it is false according to all
specifications. In all other cases, the sentence is assigned no determinate
truth-value. Supervaluationism, however, is only one way of taking
into account the verdicts of different specifications -- in technical
terms, ``aggregating'' those verdicts. In this paper, we introduce
insights from the theory of judgment aggregation to study a broad
class of different aggregation functions. This allows us to offer
an axiomatic characterization of supervaluationism, which, in turn,
locates it precisely within a larger logical space of possible approaches
to vagueness and indeterminacy. Among other results, we axiomatically
compare supervaluationism with subvaluationism, the epistemic theory
of vagueness, and Strong Kleene and LP logics.
\end{abstract}

\section{Introduction}

Whenever the interpretation of language is subject to vagueness, ambiguity,
or indeterminacy, we are faced with conflicts between several competing
assignments of truth-values to sentences. Here are some examples:\smallskip{}

\textbf{Vagueness:} If sentences such as the ``the king is bald''
or ``this amount of sugar forms a heap'' involve vague predicates
such as ``bald'' or ``heap'', then those sentences may be assigned
different truth-values, depending on where we draw the line between
``bald'' and ``not bald'' or between ``heap'' and ``no heap''.
\smallskip{}

\textbf{Ambiguity:} If legal discourse involves ambiguous terms, the
truth-values of certain sentences may depend on how we disambiguate
them. An illustration comes from a 19th-century court case in New
York City, where the point of contention was whether whale oil is
a form of fish oil. The plaintiff, the inspector of fish oil, alleged
that the defendant, a fisherman, owed a payment of \$75 for ``gauging,
inspecting and branding'' three casks of whale oil (Sainsbury 2014,
reporting on Burnett 2007). The court ruled in favour of the plaintiff,
invoking a colloquial interpretation under which ``sea-creatures
are fish'' (Sainsbury 2014), so that whale products are fish products.
The defendant insisted on a more scientific interpretation according
to which whales are mammals, not fish. 

\smallskip{}

\textbf{Indeterminacy:} Consider moral discourse in which the truth
may seem indeterminate, insofar as different criteria of ``right''
and ``wrong'' yield different verdicts. While practically all moral
theories agree that we must not torture babies and that killing innocent
people is wrong, there is much disagreement on what to say in complicated
cases, such as ``trolley problems'', in which consequentialist and
deontological theories come apart. It is here unclear what counts
as \emph{the} moral truth, all things considered.\smallskip{}

What all of these cases of vagueness, ambiguity, or indeterminacy
have in common is that, in each of them, 
\begin{lyxlist}{(iii)}
\item [{(i)}] we are considering a set of sentences, 
\item [{(ii)}] there are multiple competing perspectives on what the truth-values
of those sentences are, and 
\item [{(iii)}] we wish to establish whether at least some of the sentences
are ``true simpliciter'', or ``true, all things considered'',
or ``overall true''.
\end{lyxlist}
One of best-known philosophical proposals on how to define such an
``overall'' notion of truth -- hereafter \emph{overall truth }or
\emph{truth simpliciter} -- is \emph{supervaluationism} (Fine 1975,
Keefe 2000, Varzi 2007). On this proposal, a sentence is true simpliciter
if and only if it is true according to \emph{all} admissible perspectives;
it is false simpliciter if and only if it is false according to \emph{all}
admissible perspectives; and, by implication, it is neither true nor
false if it is true according to some admissible perspectives and
false according to others. This permits truth-value gaps, a departure
from the classical notion of truth. For instance, if the two interpretations
of ``fish'' are both admissible, then it is neither true nor false
that whale oil is fish oil. On the other hand, it is true simpliciter
that we must not torture babies, assuming that no admissible moral
theory disagrees with this verdict.

Supervaluationism has a closely related cousin, called \emph{subvaluationism}
(Ja\'{s}kowski 1969, Hyde 1997, Varzi 1997, Cobreros, Egré, Ripley,
and van Rooij 2012). According to it, a sentence is true simpliciter
if and only if it is true according to \emph{at least one} admissible
perspective; it is false simpliciter if and only if it is false according
to \emph{at least one} admissible perspective; and, by implication,
it is both true and false if it is true according to some admissible
perspectives and false according to others. While supervaluationism
gives rise to truth-value gaps, subvaluationism gives rise to truth-value
gluts: some sentences can be true and false at the same time. Whale
oil, for instance, both is and is not fish oil. In a moral dilemma,
an action may be both right and wrong. There is a vast philosophical
literature on the merits and demerits of these proposals (for a survey,
see Sorensen 2023).

In this paper, we present a general framework for capturing the relationship
between ``perspectival truth'' -- truth according to several different
perspectives -- and ``truth simpliciter''. We suggest that this
relationship can be studied using the tools of judgment-aggregation
theory (List and Pettit 2002, Dietrich 2007, Dietrich and List 2026).
Formally, truth simpliciter can be defined as a function of truth
according to the different perspectives. Using that framework, we
show that the literature's focus on supervaluationism and its kin
is not \emph{ad hoc}, but that supervaluationism and subvaluationism
stand out as the unique solutions to our problem of ``cross-perspective
aggregation'' which satisfy certain intuitively undemanding conditions.
We thereby provide axiomatic characterizations of each of these prominent
proposals. For comparison, we also provide a characterization of an
aggregation function that captures the epistemic theory of vagueness.
Finally, we show that, once we relax some of the conditions, a larger
space of possible solutions opens up, some of which have not yet been
recognized in the literature. Our framework thus provides a novel
toolbox for thinking about how perspectival truth relates to truth
simpliciter. The idea that ``{[}l{]}inguistic vagueness results from
aggregating many judgments into one'' also appears in Grinsell (2019),
but it has not yet been developed in terms of the fully general framework
of judgment-aggregation theory (Grinsell refers to standard social
choice theory in the tradition of Kenneth Arrow's 1951/1963 models
of preference aggregation).

The problem we are discussing is very general, because disagreements
between different perspectives occur in many different areas, from
language and metaphysics to morality and aesthetics, and the sources
of those disagreements can be manifold: they can be \emph{semantic}
and/or \emph{hermeneutic} (due to underspecified or competing meanings
or interpretations of language), \emph{epistemic} (due to incomplete
or conflicting information), or even \emph{ontic} (due to partial
or plural facts of the matter). While cross-perspective aggregation
problems have different interpretations depending on the sources of
the perspectival disagreement, they share a common structure, and
it is this structure that this paper is about.

\section{The aggregation of truth-evaluations}

The framework to be presented is a notational variant of the standard
framework of judgment aggregation (as, e.g., in List and Pettit 2002
and Dietrich 2007; for an overview, see Dietrich and List 2026), adapted
to the present context and suitably reinterpreted.

\subsection{The objects of evaluation}

Let $X$ be a finite set of sentences from some natural or formal
language $\mathbf{L}$ to which we wish to assign truth-values. The
set $X$ captures the domain of discourse that is to be truth-evaluated.
We assume that the underlying language can express negation: for any
sentence $\phi$ in $\mathbf{L}$, we write $\neg\phi$ for not-$\phi$.
We also assume that we have at our disposal classically well-behaved
notions of logical consistency and logical consequence.\footnote{Formally, the notion of \emph{consistency} specifies, for each subset
$Y$ of $\mathbf{L}$, whether $Y$ is \emph{consistent} or \emph{inconsistent}.
The \emph{entailment relation} specifies, for any subset $Y$ of $\mathbf{L}$
and any sentence $\phi$ in $\mathbf{L}$, whether $Y$ \emph{entails}
$\phi$ or not. To capture classical notions of consistency and entailment,
we make the following assumptions: (i) every sentence-negation pair
$\{\phi,\neg\phi\}\subseteq\mathbf{L}$ is inconsistent; (ii) subsets
of consistent sets are consistent; (iii) the empty set is consistent,
and every consistent set $S\subseteq\mathbf{L}$ has a consistent
superset $S'\subseteq\mathbf{L}$ that contains a member of each sentence-negation
pair $\{\phi,\neg\phi\}\subseteq\mathbf{L}$; and (iv) a set $S\subseteq\mathbf{L}$
entails a sentence $\phi\in\mathbf{L}$ if and only if $S\cup\left\{ \neg\phi\right\} $
is inconsistent. Call a sentence $\phi\in\mathbf{L}$ \emph{contradictory}
if $\{\phi\}$ is inconsistent, \emph{tautological} if $\{\neg\phi\}$
is inconsistent, and \emph{contingent} otherwise. For further details
on the logical underpinnings, see Dietrich (2007).} Classical propositional, predicate, modal, and conditional languages
all satisfy these conditions and could thus serve as instances of
$\mathbf{L}$, as could certain fragments of natural language. We
further assume that the set $X$ of sentences of interest contains
more than one contingent sentence-negation pair and has the structure
of a Boolean algebra, i.e., it is closed under negation ($\neg$)
and conjunction ($\wedge$), as made precise in the appendix. A natural
case is the one where $X$ is all of $\mathbf{L}$. (Several of our
results also hold under less demanding assumptions on $X$, as also
explained in the appendix.) For simplicity, we assume that double
negations cancel each other out, so we always identify $\neg\neg\phi$
with $\phi$.\footnote{The cancellation of double-negation can be made technically precise,
but the details are not relevant to our discussion.}

\subsection{Truth-evaluations}

In our framework, a \emph{truth-evaluation} of $X$ is given by a
subset $T\subseteq X$ of those sentences in $X$ that are deemed
\emph{true}. The following interpretation is natural:
\begin{itemize}
\item $\phi\in T$ means that $\phi$ is assigned the truth-value ``true''; 
\item $\neg\phi\in T$ means that $\phi$ is assigned the truth-value ``false'';
\item $\phi\notin T$ and $\neg\phi\notin T$ (if this ever happens) means
that $\phi$ is assigned no truth-value: a \emph{truth-value gap }(``neither'').
\item $\phi\in T$ and $\neg\phi\in T$ (if this ever happens) means that
$\phi$ is assigned both truth-values: a \emph{truth-value glut }(``both'').
\end{itemize}
Clearly, if the truth-evaluation behaves as assumed in classical logic,
the last two cases cannot occur: not only will all the assignments
of truth-values be consistent with one another, but there won't be
any truth-value gaps or gluts. Formally, a \emph{truth-evaluation}
$T$ is called
\begin{itemize}
\item \emph{consistent} if $T$ is a consistent set of sentences within
$X$; 
\item \emph{non-gappy} if $T$ contains at least one member of each pair
$\{\phi,\neg\phi\}\subseteq X$; 
\item \emph{non-glutty} if $T$ contains at most one member of each pair
$\{\phi,\neg\phi\}\subseteq X$; 
\item \emph{consequence-closed} if, whenever $T$ entails (in the classical
sense defined earlier) some $\phi\in X$, then $\phi\in T$ (``any
sentence entailed by true sentences is also true'').
\item \emph{rejection-closed} if, whenever $\{\neg\psi\in X:\psi\notin T\}$
entails some $\neg\phi\in X$, then $\phi\notin T$ (``any sentence
whose negation is entailed by the negations of sentences that are
not true is also not true'').
\end{itemize}
A \emph{classical} truth-evaluation is one with all five properties.
It corresponds to a truth-function, as standardly defined, for the
set $X$. Note the following relations between these properties:

\begin{remark}For any truth-evaluation $T$, the following are equivalent:
(i) $T$ is classical; (ii) $T$ is non-gappy and consistent; (iii)
$T$ is non-gappy, non-glutty, and consequence-closed; (iv) $T$ is
non-gappy, non-glutty, and rejection-closed.\end{remark}

\subsection{The notion of aggregation}

Suppose now we are given multiple truth-evaluations and wish to ``aggregate''
them into an ``overall'' truth-evaluation. We speak of truth-evaluations
``from different perspectives'', where this is intended to be neutral
on how these ``perspectives'' are to be understood and where they
come from. For instance, we remain neutral on whether different perspectives
should be understood as expressing different \emph{judgments} about
the truth or as different \emph{criteria} or \emph{standards} as to
what the truth is; our formal analysis applies across these interpretive
differences. One particularly relevant way to interpret ``aggregation''
here is in terms of \emph{supervenience}: the overall truth-evaluation
supervenes on the given collection of truth-evaluations from different
perspectives, meaning that it is ``metaphysically determined'' by
them. Roughly speaking, it is impossible for there to be a difference
at the level of overall truth without some difference at the level
of the underlying perspectival truths.

Let us label the perspectives $N=\{1,2,...,n\}$, where $n\geq2$.
For the moment, there are finitely many perspectives; later, we lift
than restriction. An $n$-tuple of truth-evaluations across them,
$\left\langle T_{1},T_{2},...,T_{n}\right\rangle $, is called a \emph{profile
of truth-evaluations}. An \emph{aggregation function} is a function
that takes as input a profile of truth-evaluations (from some domain
of admissible profiles) and produces as output a single truth-evaluation
$T$: the \textquotedblleft overall\textquotedblright{} truth-evaluation.
For the sake of generality, we do not build into the definition of
an aggregation function any further assumptions about the domain of
admissible inputs and the co-domain of admissible outputs. We deliberately
leave open, in particular, whether the truth-evaluations on the input
and output sides must be classical. We introduce appropriate restrictions
later.

\subsection{Examples}

Examples of aggregation functions, which are each defined for every
profile $\left\langle T_{1},T_{2},...,T_{n}\right\rangle $ of classical
truth-evaluations, are: 
\begin{itemize}
\item the\emph{ majority function}: $T=\{$$\phi\in X:|\{i\in N:\phi\in T_{i}\}|>n/2\}$;
\item a \emph{quota function} (which subsumes super-, simple, and sub-majority
functions): $T=\{$$\phi\in X:\frac{|\{i\in N:\phi\in T_{i}\}|}{n}\geq q\}$,
where $q$ is some antecedently fixed ``acceptance quota'' between
0 and 1;
\item a \emph{dictatorship} of one perspective: $T=T_{i}$, where $i$ is
some antecedently fixed perspective in $N$;
\item \emph{supervaluationism}: $T=\underset{i\in N}{\bigcap}T_{i}$;
\item \emph{subvaluationism}: $T=\underset{i\in N}{\bigcup}T_{i}$.
\end{itemize}
Table 1 gives an example of how some of these different aggregation
functions behave. It should be evident that they all have some problems.
The majority function may sometimes generate inconsistent overall
truth-evaluations even when the input truth-evaluations are each consistent,
for instance by deeming each of $\phi$ and $\psi$ true and yet $\phi\wedge\psi$
false. Other quota functions (certainly sub-majority functions but
even super-majority functions) can generate similar problems, depending
on the quota; we return to super- and submajority functions later.
Dictatorships pay attention only to one \textquotedblleft privileged\textquotedblright{}
perspective (e.g., perspective 1 in Table 1) and completely ignore
all other perspectives, though we will later relate it to the epistemic
theory of vagueness. Supervaluationism may, and typically does, generate
gappy truth-evaluations, and subvaluationism may, and typically does,
generate glutty truth-evaluations, as also illustrated in Table 1.
In what follows, however, we show that there is a precise formal sense
in which supervaluationism, subvaluationism, and dictatorships stand
out.
\begin{table}[H]
\centering%
\begin{tabular}{|c|c|c|c|c|c|c|}
\hline 
 & $\phi$ & $\psi$ & $\phi\wedge\psi$ & $\neg\phi$ & $\phi\vee\neg\phi$ & $\phi\wedge\neg\phi$\tabularnewline
\hline 
\hline 
Perspective 1 & True & True & True & False & True & False\tabularnewline
\hline 
Perspective 2 & True & False & False & False & True & False\tabularnewline
\hline 
Perspective 3 & False & True & False & True & True & False\tabularnewline
\hline 
Majority & True & True & False & False & True & False\tabularnewline
\hline 
\begin{cellvarwidth}[t]
\centering
Dictatorship of

perspective 1
\end{cellvarwidth} & True & True & True & False & True & False\tabularnewline
\hline 
Supervaluationism & Neither & Neither & Neither & Neither & True & False\tabularnewline
\hline 
Subvaluationism & Both & Both & Both & Both & True & False\tabularnewline
\hline 
\end{tabular}\caption{Different aggregation functions}
\end{table}

\section{Three axiomatic characterization results}

In line with the standard approach in the theory of judgment aggregation,
we do not merely select an aggregation function \emph{ad hoc}, but
introduce some conditions that we might expect a plausible aggregation
function to meet and ask which aggregation functions, if any, meet
those conditions. 

\subsection{A characterization of supervaluationism}

We begin with five conditions that may be viewed as plausible constraints
on the relationship between perspectival truth and ``overall'' truth.
They are best viewed as baseline conditions. We explore their implications
but subsequently also discuss how, and when, they may be relaxed.
The first condition requires that the aggregation function be well-defined
for every profile of classical truth-evaluations of $X$. (The background
assumption is that each of the competing perspectives individually
yields a classical truth-evaluation.)

\bigskip{}

\noindent\textbf{Unrestricted domain}. \sloppy The aggregation function
admits as input every profile $\left\langle T_{1},T_{2},...,T_{n}\right\rangle $
of classical truth-evaluations.

\bigskip{}

\noindent This condition seems reasonable insofar as the sentences
in $X$ may express contingent matters whose truth-value may vary
across different possible situations. Our aggregation function should
thus remain well-defined across such possible variation.

The second condition requires that ``overall'' truth, as specified
by the aggregation function, be consequence-closed.

\bigskip{}

\noindent\textbf{Consequence-closed truth}. The output $T$ generated
by the aggregation function is always a consequence-closed truth-evaluation. 

\bigskip{}

\noindent This is of course a standard condition on truth. The third
condition requires that, in the special case in which \emph{all} perspectives
agree on the truth-values of \emph{every} sentence in $X$ (i.e.,
there is no disagreement whatsoever), this agreed truth-evaluation
become the overall evaluation $T$.

\bigskip{}

\noindent\textbf{Global unanimity preservation}. For any profile
$\left\langle T_{1},T_{2},...,T_{n}\right\rangle $ in the domain,
if $T_{1}=T_{2}=...=T_{n}$, then $T=T_{1}=T_{2}=...=T_{n}$.\footnote{\noindent In the presence of unrestricted domain and sentence-wise
independence (as defined below), global unanimity preservation implies
the following ``local'' unanimity-preservation requirement: for
any profile $\left\langle T_{1},T_{2},...,T_{n}\right\rangle $ in
the domain and any $\phi\in X$, if $\phi\in T_{i}$ for all $i\in N$,
then $\phi\in T$.}

\bigskip{}

\noindent It would be strange if, in the case of global agreement
across all perspectives -- i.e., agreement with respect to the entire
domain of discourse -- we did not take this to settle the overall
assignment of truth-values. The fourth condition requires that the
overall truth-value of each sentence depend only on the different
perspectives' truth-values for that sentence, not on the truth-values
they assign to other sentences.

\bigskip{}

\noindent\textbf{Sentence-wise independence}. For any profiles $\left\langle T_{1},T_{2},...,T_{n}\right\rangle $
and $\left\langle T_{1}^{*},T_{2}^{*},...,T_{n}^{*}\right\rangle $
in the domain and any $\phi\in X$, if $\left[\phi\in T_{i}\Leftrightarrow\phi\in T_{i}^{*}\mbox{ for all }i\in N\right]$
then $\left[\phi\in T\Leftrightarrow\phi\in T^{*}\right]$.

\bigskip{}

\noindent We can think of this as a condition of \emph{local} or \emph{sentence-wise}
supervenience between perspectival truth and overall truth. The final
condition requires that no perspective be totally irrelevant, in the
sense that, for each perspective, it is possible, in some circumstances,
to make a difference to the overall truth-evaluation, however small
that difference might be. \bigskip{}

\noindent\textbf{Sensitivity.} For every perspective $i$, there
exists at least one pair of profiles $\left\langle T_{1},...,T_{n}\right\rangle $
and $\left\langle T_{1},...,T_{i}^{*},...,T_{n}\right\rangle $ which
differ only in the truth-evaluation for perspective $i$ (i.e., $T_{i}\neq T_{i}^{*}$)
such that the resulting overall truth-evaluations $T$ and $T^{*}$
are different. \bigskip{}

Surprisingly, the five conditions jointly characterize supervaluationism.
This is a variant of a result in the theory of judgment aggregation
(Dietrich and List 2008, Dokow and Holzman 2010b); we provide a proof
sketch later.

\begin{theorem}Supervaluationism is the only aggregation function
satisfying unrestricted domain, consequence-closed truth, global unanimity
preservation, sentence-wise independence, and sensitivity.\end{theorem}

There is no other way of satisfying these conditions. Furthermore,
since supervaluationism generates not only consequence-closed but
also consistent truth-evaluations, we might add a further condition
to our list, though it is already entailed by the other conditions:\bigskip{}

\noindent\textbf{Consistent truth}. The output $T$ generated by
the aggregation function is always a consistent truth-evaluation. 

\bigskip{}

By contrast, supervaluationism violates the following condition, as
already illustrated in Table 1:\bigskip{}

\noindent\textbf{No truth-value gaps}. The output $T$ generated
by the aggregation function is never gappy.

\bigskip{}

Evidently, under supervaluationism, the all-things-considered truth-evaluation
$T$ will be gappy with respect to every sentence-negation pair on
which the different truth-evaluations $T_{1},T_{2},...,T_{n}$ do
not unanimously agree.

\subsection{A characterization of subvaluationism}

We will now see that a slightly modified list of conditions will lead
to subvaluationism rather than supervaluationism. Suppose, again,
we would like our aggregation function to satisfy the conditions of
unrestricted domain, global unanimity preservation, sentence-wise
independence, and sensitivity. But suppose we replace the condition
of consequence-closed truth with the condition of rejection-closed
truth:

\bigskip{}

\noindent\textbf{Rejection-closed truth}. The output $T$ generated
by the aggregation function is always a rejection-closed truth-evaluation. 

\bigskip{}

The resulting five conditions now characterize subvaluationism, a
result that can be viewed as the logical ``dual'' of the earlier
theorem: 

\begin{theorem}Subvaluationism is the only aggregation function satisfying
unrestricted domain, rejection-closed truth, global unanimity preservation,
sentence-wise independence, and sensitivity.\end{theorem}

Again, there is no other way of satisfying these conditions. We also
sketch the proof later. Notice that while supervaluationism fails
to satisfy the requirement of no truth-value gaps, subvaluationism
correspondingly fails to satisfy the requirement of no truth-value
gluts:

\bigskip{}

\noindent\textbf{No truth-value gluts}. The output $T$ generated
by the aggregation function is never glutty.

\bigskip{}

\noindent Again, recall Table 1. Clearly, subvaluationism may generate
logically inconsistent truth-evaluations, for instance if a sentence
(e.g., ``whale oil is fish oil'') and its negation (e.g., ``whale
oil is not fish oil'') are both deemed to be true. However, subvaluationism
satisfies a natural ``dual'' of consistency. Call a truth-evaluation
$T$ \emph{rejection-consistent} if the set of negations of those
sentences in $X$ that are not deemed true by $T$ is consistent,
formally if $\{\neg\psi\in X:\psi\notin T\}$ is a consistent set.
Then subvaluationism satisfies the following:

\bigskip{}

\noindent\textbf{Rejection-consistent truth}. The output $T$ generated
by the aggregation function is always a rejection-consistent truth-evaluation. 

\subsection{A characterization of a dictatorship of one perspective}

Suppose now we would like overall truth to be not merely consequence-closed
but classical, especially \emph{bivalent}: there should be no truth-value
gaps or truth-value gluts.\bigskip{}

\noindent\textbf{Classical truth}. The output $T$ generated by the
aggregation function is always a classical truth-evaluation. 

\bigskip{}

If we keep in place our other conditions, we immediately get an impossibility
result:

\begin{theorem}There are no aggregation functions satisfying unrestricted
domain, classical truth, global unanimity preservation, sentence-wise
independence, and sensitivity.\end{theorem}

The reason is that classical truth implies both consequence-closedness
and rejection-closedness, and, in the context of the other conditions,
the one condition implies supervaluationism while the other implies
subvaluationism. Since there is more than one perspective (recall
that $n\geq2$), we cannot have both. 

One salient way to avoid the impossibility -- among others, which
will be discussed later -- is to drop the condition of sensitivity,
and to allow that not all perspectives matter. But it then turns out
that only a single perspective matters. This follows from another
result from judgment aggregation theory (Dietrich and List 2007a,
Dokow and Holzman 2010a):

\begin{theorem}An aggregation function satisfies unrestricted domain,
classical truth, global unanimity preservation, and sentence-wise
independence if and only if it is a dictatorship of one perspective.\end{theorem}

Once more, there is no other way of satisfying these conditions. If
the relation between perspectival truth and overall truth is to satisfy
the theorem's four conditions, then there must be a ``dictatorial''
perspective. In other words, one perspective must be deemed to be
the ``correct'' one, so that overall truth is simply truth from
that single correct perspective. 

This result is directly relevant to the epistemic theory of vagueness.
According to this theory, prominently defended by Williamson (1994),
there is a fact of the matter as to which of several seemingly admissible
perspectives is the true correct one, even though we may be ignorant
of this fact. In the case of vague language, the different perspectives
may stem from different precisifications of certain vague predicates,
corresponding to different cut-off lines between ``heap'' and ``no
heap'' or between ``bald'' and ``not bald'', for example. But
while we may be ignorant about where to draw the line, and therefore
ignorant about which perspective is the correct one, there is a fact
of the matter. Our theorem nicely characterizes this view. Suppose
there are different seemingly admissible perspectives, and these jointly
suffice to determine the overall truth, in the sense that overall
truth supervenes on those perspectives (or on some of them). And suppose
further that the supervenience relation satisfies unrestricted domain
(i.e., overall truth is always well defined, no matter what the different
perspectives say), global unanimity preservation (i.e., if all the
perspectives agree on the entire domain of discourse, then the overall
truth coincides with that unanimously agreed truth), and sentence-wise
independence (i.e., the overall truth-status of any sentence supervenes
on the perspectival truth-status of that sentence). Then, if overall
truth is classical, there must exist a ``dictatorial'' perspective,
i.e., one perspective must be the true correct one, even if we don't
know which one it is. Consistently with the epistemic theory of vagueness,
the present theorem's conditions only imply the \emph{existence} of
some dictatorial perspective; they don't answer the question of which
one it is.

Theorem 4 was originally formulated to address the question of how
several individuals' judgments on some logically related propositions
-- say, in a court or in an expert panel -- can be aggregated into
collective judgments, with the surprising answer that under certain
conditions only a dictatorial aggregation function will work. Interestingly,
this also yields the most famous result of social choice theory as
a corollary, namely Arrow's classic impossibility theorem on the aggregation
of preferences, according to which there exists no non-dictatorial
function for aggregating several individuals' preferences over three
or more options into collective preferences in accordance with some
minimal conditions. To see the formal connection between preferences
and judgments, note that preferences can be \emph{formally} represented
as ``judgments'' on binary comparisons of the form ``A is better
than B'', ``B is better than C'', etc. For instance, if someone
prefers A to B to C, then -- setting aside interpretational differences
-- this can be formally represented as a ``judgment'' that A is
better than B, that B is better than C, and that A is better than
C. Thus, somewhat surprisingly, our present characterization of the
epistemic theory of vagueness and Arrow's famous impossibility theorem
(1951/1963) can be subsumed under a common formal structure: they
both derive from what may be described as ``Arrow's theorem in judgment
aggregation''.

\section{A proof sketch}

Why do the present characterization results hold? It is instructive
to sketch the basic proof strategy. Let us begin with the characterization
of supervaluationism. Consider an aggregation function that satisfies
unrestricted domain, consequence-closed truth, global unanimity preservation,
and sentence-wise independence. For the moment, let's not require
sensitivity. The first thing to note is that because of sentence-wise
independence, the overall truth-status of any sentence $\phi$ depends
only on its truth-status according to each of the given perspectives,
and so the aggregation function can be formally represented as follows:

\bigskip{}

\noindent\textbf{Winning coalition representation.} For any admissible
profile $\left\langle T_{1},T_{2},...,T_{n}\right\rangle $, $T$
consists of all and only those sentences $\phi$ in $X$ for which
the set of perspectives deeming $\phi$ true is a ``winning coalition''
for $\phi$. 

\bigskip{}

A \emph{winning coalition} for $\phi$ is a subset $M$ of $N$ such
that, for every admissible profile $\left\langle T_{1},T_{2},...,T_{n}\right\rangle $,
\[
\text{if }\phi\in T_{i}\text{ for all }i\in M\text{ and }\phi\notin T_{i}\text{ for all }i\notin M,\text{ then }\phi\in T.
\]
When a set of perspectives is a winning coalition for $\phi$, a scenario
in which these perspectives agree that $\phi$ is true while all other
perspectives disagree suffices for the overall verdict that $\phi$
is true. It is only because the aggregation function satisfies sentence-wise
independence that we can have this winning-coalition representation
of it. Our characterization of supervaluationism then proceeds as
follows. 

First of all, the Boolean structure of the set $X$ of sentences under
consideration has a key implication: any winning coalition for any
sentence $\phi$ in $X$ must also be a winning coalition for any
other sentence $\psi$ in $X$; otherwise our aggregation function
might fail to produce consequence-closed truth-evaluations. The aggregation
function can then be represented in terms of a single formal ``parameter'',
namely a single specification of which subsets of $N$ are winning
coalitions. Call the collection of all winning coalitions $\mathcal{W}$.
This is the collection of all those sets of perspectives -- subsets
of $N$ -- that each form a winning coalition for any sentence $\phi$
in $X$. By specifying $\mathcal{W}$, we have fully specified the
aggregation function.

The next step is to show that this collection of winning coalitions
$\mathcal{W}$ is a \emph{filter} on the set of perspectives, i.e.,
a set of subsets of $N$ that satisfies
\begin{lyxlist}{(iii)}
\item [{(i)}] \emph{non-triviality}: $N$ is contained in $\mathcal{W}$,
while the empty set is not;
\item [{(ii)}] \emph{superset closedness}: if $M$ is contained in $\mathcal{W}$,
then so is any superset $M'$ of $M$ (with $M'\subseteq N$);
\item [{(iii)}] \emph{intersection closedness}: if $M$ and $M'$ are each
contained in $\mathcal{W}$, then so is their intersection $M\cap M'$.
\end{lyxlist}
Non-triviality follows from global unanimity preservation, together
with unrestricted domain. Superset closedness follows from the fact
that, under the given conditions, the aggregation function can be
shown to be monotonic, meaning that if a particular set $M$ of perspectives,
by agreeing that $\phi$ is true (while all other perspectives disagree),
suffices to bring about the verdict that $\phi$ is overall true,
then so does any superset of $M$. Intersection closedness, finally,
can be derived from the Boolean structure of $X$ and the requirement
that $T$ be consequence-closed.

Since the collection $\mathcal{W}$ of all winning coalitions is a
filter on $N$, and $N$ is a finite set, the intersection of all
winning coalitions must be a winning coalition itself; call it $M$.
Moreover, since the empty set is not a winning coalition, $M$ is
non-empty. It then follows that there exists a set of perspectives,
namely $M$, such that the all-things considered truth is whatever
all the perspectives in $M$ agree on, formally, for any admissible
profile $\left\langle T_{1},T_{2},...,T_{n}\right\rangle $, we have
$T=\underset{i\in M}{\bigcap}T_{i}$. This establishes that the aggregation
function is what is called ``oligarchic'' (Dietrich and List 2008,
Dokow and Holzman 2010b). To achieve sensitivity as well, no perspective
must be excluded from $M$, and so $M$ must be $N$ in its entirety.
This yields supervaluationism.

To characterize subvaluationism, we proceed in the same way, but just
with one modification. Replacing consequence-closed truth with rejection-closed
truth implies that, this time, it is not the collection of \emph{winning
coalitions} that is a filter on $N$ but the collection of \emph{negatively}
\emph{winning coalitions}. A \emph{negatively winning coalition} for
$\phi$ is a subset $M$ of $N$ such that, for every admissible profile
$\left\langle T_{1},T_{2},...,T_{n}\right\rangle $, 
\[
\text{if }\phi\notin T_{i}\text{ for all }i\in M\text{ and }\phi\in T_{i}\text{ for all }i\notin M,\text{ then }\phi\notin T.
\]
When a set of perspectives is negatively winning for $\phi$, these
perspectives, by agreeing that $\phi$ is not true while all other
perspectives disagree, will jointly bring about the overall verdict
that $\phi$ is not true. The collection of all negatively winning
coalitions can be shown to be non-trivial, superset-closed, and intersection-closed.
The intersection of all negatively winning coalitions will then be
non-empty and negatively winning as well; call it $M$. To achieve
sensitivity, $M$ must again coincide with all of $N$. In fact, $N$
is the only negatively winning coalition. Since the complement of
any set of perspectives that is not negatively winning must be winning
in the ordinary positive sense, it follows that the only set of perspectives
that is \emph{not} winning is the empty set. Every non-empty subset
of $N$ must then be a winning coalition. So, formally, for any admissible
profile $\left\langle T_{1},T_{2},...,T_{n}\right\rangle $, we have
$T=\underset{i\in N}{\bigcup}T_{i}$. This yields subvaluationism. 

Finally, to characterize a dictatorship of one perspective, suppose
now that our aggregation function must satisfy classical truth: the
overall truth-evaluation $T$ must always be classical, which rules
out both truth-value gaps and truth-value gluts. To avoid an outright
impossibility (since no aggregation function can be simultaneously
supervaluationist and subvaluationist), we do not require sensitivity.
So, we are thereby permitting that not all perspectives make a difference.
Consider again the collection of winning coalitions, in the original
positive sense. To avoid truth-value gaps and truth-value gluts, this
collection must satisfy completeness:
\begin{lyxlist}{(iii)}
\item [{(iv)}] \emph{completeness}: for any subset $M$ of $N$, either
$M$ or its complement $N\backslash M$ is contained in $\mathcal{W}$.
\end{lyxlist}
However, a collection $\mathcal{W}$ of winning coalitions that satisfies
not just conditions (i) to (iii) but also condition (iv) is not merely
a \emph{filter} on $N$ but an \emph{ultrafilter}. And since $N$
is finite, any ultrafilter is \emph{principal}, i.e., there exists
a single element $i$ of $N$ which belongs to all members of $\mathcal{W}$,
i.e., which belongs to every winning coalition. This implies that
any sentence is deemed overall true if and only if it is deemed true
by perspective $i$. We then have a dictatorship of perspective $i$.

\section{Escape routes}

As we have seen, unrestricted domain, consequence- or rejection-closed
truth, global unanimity preservation, sentence-wise independence,
and sensitivity jointly imply that the relationship between perspectival
truth and overall truth must have a supervaluationist or subvaluationist
structure. If we further require truth to be classical, we face an
outright impossibility result: no aggregation function will jointly
satisfy our conditions. If we are then willing to abandon sensitivity,
we must opt for a dictatorship of one perspective, consistently with
the epistemic theory of vagueness. Those who wish to avoid these conclusions
must relax some of the conditions of our characterization theorems.
Let us briefly discuss the options. We will run through them in reverse
order.

\subsection{Relaxing sensitivity}

Recall that sensitivity requires that no perspective be totally irrelevant
in the determination of overall truth. If we do not require this,
then, as our proof shows, the aggregation function must have an ``oligarchic''
structure: there must exist some non-empty subset $M$ of the set
$N$, which need not coincide with $N$, such that, for any admissible
profile $\left\langle T_{1},T_{2},...,T_{n}\right\rangle $, 
\begin{itemize}
\item $T=\underset{i\in M}{\bigcap}T_{i}$, if we demand consequence-closed
truth;
\item $T=\underset{i\in M}{\bigcup}T_{i}$, if we demand rejection-closed
truth; and
\item $T=T_{i}$ with $M=\left\{ i\right\} $ being a singleton set, if
we demand classical truth.
\end{itemize}
We can view the first possibility as a generalized version of supervaluationism,
where only the perspectives in $M$ are deemed relevant, and the second
possibility as a generalized version of subvaluationism, where again
only the perspectives in $M$ are relevant. The last possibility was
already covered by our third characterization theorem. These escape
routes from our results are least interesting, since they are structurally
not very different from the possibilities already identified by our
main results.

\subsection{Relaxing sentence-wise independence}

Recall that sentence-wise independence requires that the ``overall''
truth-value of each sentence depend only on the different perspectives'
truth-values for that sentence, not on the truth-values assigned to
other sentences. We can think of this as a requirement of \emph{local}
supervenience: the overall truth-status of each sentence supervenes
on the truth-status \emph{of that sentence} according to the difference
perspectives; those perspectives' verdicts on other sentences are
irrelevant. One might worry, in particular, that this goes against
the spirit of the widely accepted compositionality requirement on
truth, namely that the truth-values of composite sentences should
be functionally determined by the truth-values of their components.
Under sentence-wise independence, when we determine the overall truth-value
of $\phi\wedge\psi$, for instance, we must consider only the truth-values
of $\phi\wedge\psi$ according to the different perspectives, not
the truth-values of each of $\phi$ and $\psi$, taken in isolation.
Although this does not formally rule out that the truth-value assigned
to the composite sentence could match the one implied by the truth-values
assigned to its components, compositionality is not ``procedurally''
built into the design of the aggregation function. An illustration
of a compositionality breach comes from supervaluationism, where $\phi\vee\neg\phi$
is assigned the truth-value ``true'' even when neither $\phi$ nor
$\neg\phi$ is deemed ``true'' (Cobreros and Tranchini 2014, Williamson
1994). Indeed, only rather special sentence-wise independent aggregation
functions yield overall truth-evaluations that respect compositionality,
most notably the dictatorial ones. If we give up sentence-wise independence,
many aggregation functions become possible. We will single out two
kinds of possibilities. The first kind comes from judgment-aggregation
theory and yields largely classical truth-evaluations. The second
kind is inspired by some non-classical logics of vagueness.

Let us begin with the aggregation functions that come directly from
the context of judgment aggregation, where a group seeks to form collective
judgments on some propositions based on its members' individual judgments.
We will simply mention those possibilities, without taking a stand
on how plausible they are as truth-determination functions, governing
the relationship between perspectival and overall truth.\footnote{For a full formal review of the aggregation-theoretic possibilities,
see Dietrich and List (2026).}
\begin{itemize}
\item \emph{Premise-based functions}: Here, a (negation-closed) subset of
the sentences in $X$ is identified as a set of ``premises'', where
this premise set contains no logical connections between distinct
sentence-negation pairs. It is then easy to assign consistent overall
truth-values to those premises. The overall truth-values of all premises
can be determined, for instance, by majority voting across the different
perspectives (i.e., each premise is assigned the truth-value assigned
to it by the majority of perspectives, perhaps with a tie-breaker
in case of majority ties). Finally, the overall truth-values of all
other sentences are determined by classical logical entailment or
left gappy in case the premise verdicts have no implications for them.
The result is a consistent and consequence-closed truth-evaluation.
The result is fully classical if the premise set is chosen as a logical
basis for the entire set $X$ (so that settling the premises suffices
to settle all sentences in $X$).
\item \emph{Priority functions}: Here, the sentences in $X$ are considered
in some given order of priority.\footnote{Technically, the priority order is an order of sentence-negation pairs
$\left\{ \phi,\neg\phi\right\} $. To say that a sentence $\phi$
is prior to another is to say that the pair to which it belongs is
prior to the pair to which the other sentence belongs. We assume that
the priority order is a linear order, so that the pairs can be ranked
from highest to lowest priority. This could be generalized (Dietrich
and List 2026).} The overall truth-value of each sentence is now determined as follows.
If that truth-value is not yet constrained (i.e., classically logically
entailed in the earlier sense) by the overall truth-values assigned
to sentences considered previously in the given order, then it is
determined by majority voting across the different perspectives. If
the truth-value is already constrained by the truth-values assigned
to previously considered sentences, then it is determined according
to the relevant entailments. If there are never any majority ties
among the perspectives, the result is a classical truth-value assignment.
\item \emph{Distance-based functions}: Here, we introduce a notion of \emph{distance}
between truth-evaluations. The simplest such notion is \emph{Hamming
distance}. Here, the distance between any two truth-evaluations $T$
and $T'$ is simply the number of sentences on which $T$ and $T'$
disagree, i.e., $d(T,T')=\left|\left\{ \phi\in X:\phi\in T\nLeftrightarrow\phi\in T'\right\} \right|$.
For any admissible profile of truth-evaluations $\left\langle T_{1},T_{2},...,T_{n}\right\rangle $,
the overall truth-evaluation is now simply one that minimizes the
sum-total distance from the perspective-specific truth-evaluations,
i.e., one for which $\sum_{i\in N}d(T,T_{i})$ is minimal (with some
tie-breaking rule if there is more than one distance-minimizing truth-evaluation)
(Pigozzi 2006).
\end{itemize}
Next let us turn to some possibilities inspired by the literature
on vagueness (e.g., Ripley 2013, Marcos, P\v{r}enosil, and Egré 2026).
Some scholars have challenged supervaluationism precisely because
it breaches compositionality (e.g., Tye 1989, Ripley 2013). Recall
that supervaluationism would say that $\phi\vee\neg\phi$ is overall
true even when neither $\phi$ nor $\neg\phi$ is overall true. One
might criticize this. To illustrate, one might question whether a
sentence such as ``the king is bald or not bald'' should be considered
overall true when the king is a borderline case of baldness, in the
sense that it is neither true that the king is bald nor true that
he is not bald. In particular, one might say: instances of the law
of excluded middle, of the form $\phi\vee\neg\phi$, are neither true
nor false when the components, $\phi$ and $\neg\phi$, are neither
true nor false. 

A logic of vagueness that seeks to capture this idea is Strong Kleene
Logic. It can be introduced as follows. Assuming that the sentences
in $X$ come from propositional logic, we first assign the truth-values
``true'', ``false'', or ``neither'' to all atomic sentences
-- those without any logical connectives -- in a way that matches
the supervaluationist assignments to them. Specifically, we can assign
the truth-values ``true'' and ``false'' in the clear-cut cases
in which all admissible perspectives agree and ``neither'' in all
cases of disagreement (``borderline'' cases). We then assign truth-values
to all non-atomic sentences -- say, those with the connectives $\neg,\wedge,\vee$
-- in line with the following compositional rules:
\begin{itemize}
\item $\neg\phi$ is ``true'' if and only if $\phi$ is ``false''; $\neg\phi$
is ``neither'' if and only if $\phi$ is ``neither'';
\item $\phi\wedge\psi$ is ``true'' if and only if both $\phi$ and $\psi$
are ``true''; $\phi\wedge\psi$ is ``false'' if and only if at
least one of $\phi$ or $\psi$ is false; $\phi\wedge\psi$ is ``neither''
in all remaining cases;
\item $\phi\vee\psi$ is ``true'' if and only if at least one of $\phi$
or $\psi$ is ``true''; $\phi\vee\psi$ is ``false'' if and only
if both of $\phi$ or $\psi$ are false; $\phi\vee\psi$ is ``neither''
in all remaining cases.
\end{itemize}
To represent this as an aggregation function, we can redefine a premise-based
aggregation function as follows. Let the ``premise set'' consist
of all atomic sentences and their negations. For each profile of truth-evaluations,
assign the overall truth-values to the premises via the usual supervaluationist
criterion. But now derive the truth-values of all other sentences,
not via classical logical entailment (as in a standard premise-based
function), but via the above-stated compositional rules: in effect,
applying a non-classical notion of consequence. The result is a non-classical
premise-based function that outputs a Strong Kleene truth-evaluation.

Just as supervaluationism has a dual in the form of subvaluationism,
so Strong Kleene Logic has a dual, which is called the Logic of Paradox
(LP) (for discussion, see Ripley 2013). Its compositional rules are
the same as the ones for Strong Kleene Logic, with truth-value gluts
replacing truth-value gaps. A redefined non-classical premise-based
function then yields LP assignments. Here, we start from a subvaluationist
assignment on the premises and use the LP compositional rules for
assigning the truth-values of all other sentences. It is thus worth
noting that those non-classical truth-value assignments (Strong Kleene
and LP) can also be recovered within our aggregation-theoretic framework.

If we opt for an aggregation function that gives up sentence-wise
independence -- whether of a classical or non-classical kind --
we retain the idea that overall truth supervenes on perspectival truth.
However, we give up the ``local'' or ``sentence-wise'' version
of this idea, according to which the truth-value of each sentence
supervenes on the truth-values assigned to that sentence by the different
perspectives. Rather, the supervenience relation can become holistic,
in the sense that the overall \emph{web of truth-values} supervenes
on the \emph{webs of truth-values} assigned by the different perspectives.
The advantage, however, is the preservation of compositionality.

\subsection{Relaxing global unanimity preservation}

Recall that global unanimity preservation simply requires that when
all the different perspectives are in complete agreement with one
another, in the sense that they agree on the truth-values of \emph{all}
sentences in $X$, then this agreed truth-evaluation becomes the overall
one. It is hard to see how we could plausibly relax this. But even
if we did, only very uninteresting aggregation functions would become
possible (see also Dietrich and List 2026). In particular, assuming
we are also willing to relax sensitivity, the main possibilities are
constant functions, according to which the overall truth-evaluation
$T$ is exogenously fixed and does not depend at all on the perspective-specific
truth-evaluations. That is: the overall truth-evaluation is the same
constant $T$ for every profile $\left\langle T_{1},T_{2},...,T_{n}\right\rangle $.
In effect, overall truth does not ``listen'' to perspective-specific
truth at all. Moreover, we would still get the following impossibility
result (a corollary of a result from Dietrich 2006):

\begin{theorem}There are no aggregation functions satisfying unrestricted
domain, classical truth, sentence-wise independence, and sensitivity.\end{theorem}

Without the requirement of sensitivity, the given conditions would
characterize the class of either dictatorial or constant functions,
neither of which satisfy sensitivity.

\subsection{Relaxing consequence- or rejection-closed truth}

Suppose we weaken the requirements on overall truth and allow that
overall truth may fail to be either consequence- or rejection-closed.
However, let us still require overall truth to be consistent: it should
not be the case that the set of sentences deemed true forms a logically
inconsistent set. We then get an elegant possibility result. To state
it, we need some preliminary definitions. 

Call a set of sentences \emph{minimal inconsistent} if it is inconsistent
but all its proper subsets are consistent. To give a simple example
(for a moment setting aside the requirement that $X$ be an entire
Boolean algebra), if $X$ consists of $\phi,\psi,\phi\wedge\psi,\phi\vee\psi$
and the negations of these sentences (all in standard propositional
logic), then $\left\{ \phi,\neg\phi\right\} $, $\left\{ \phi,\psi,\neg(\phi\wedge\psi)\right\} $,
and $\left\{ \neg\phi,\neg\psi,\phi\vee\psi\right\} $ are all instances
of minimal inconsistent sets. Now let $k$ be the size of a \emph{largest}
minimal inconsistent subset of $X$. In the present example, $\left\{ \phi,\psi,\neg(\phi\wedge\psi)\right\} $,
and $\left\{ \neg\phi,\neg\psi,\phi\vee\psi\right\} $ are the two
largest minimal inconsistent subsets of $X$, so $k=3$. We can think
of $k$ as a simple measure of the complexity of the logical interconnections
among the sentences in $X$. The following result holds (from Dietrich
and List 2007b):\footnote{For an earlier related discussion of using notions such as minimal
inconsistency to measure the complexity of logical interconnections,
compare also Knight (2002).}

\begin{theorem}A\emph{ }supermajority function with an acceptance
quota above $\frac{k-1}{k}$ guarantees a consistent overall truth-evaluation,
where $k$ is the size of a largest minimal inconsistent subset of
$X$.\end{theorem}

Why does this hold? For $T$ to be inconsistent, it would have to
include some minimal inconsistent set of sentences; after all, every
finite inconsistent set has at least one minimal inconsistent subset.
This would contain at most $k$ sentences, because $k$ is the size
of a largest minimal inconsistent subset of $X$. These sentences
would each be deemed true by a supermajority of more than $\frac{k-1}{k}$
among the $n$ perspectives. But it is easy to see that any $k$ or
fewer supermajorities of size above $\frac{k-1}{k}$ must have some
non-empty overlap: any two distinct simple majorities (above one half)
must overlap, any three distinct supermajorities of more than two
thirds must overlap, and so on. This means that at least one perspective
must deem all these $k$ or fewer sentences true. But because these
sentences are mutually inconsistent, this contradicts the assumption
that each perspective yields a classical truth-evaluation, in particular
a consistent one. So, $T$ is consistent, after all. What we get,
via the present supermajority function, is a consistent but not generally
consequence-closed truth-evaluation. Only the limiting case in which
the acceptance quota is pushed up to 100\% is consequence-closed.
This limiting case is supervaluationism.

Interestingly, we can also identify a ``dual'' possibility. Starting
from the baseline case of subvaluationism, we can ask what happens
if we relax rejection-closedness. Again, let $k$ be defined the size
of a largest minimal inconsistent subset of $X$. Here, a submajority
function with an acceptance quota of $\frac{1}{k}$ is the analogue
of the supermajority function of the previous result. Such an aggregation
function deems any sentence true if it is true according to at least
$\frac{1}{k}$ among the $n$ perspectives.

\begin{theorem}A\emph{ }submajority function with an acceptance quota
no more demanding than $\frac{1}{k}$ guarantees a rejection-consistent
overall truth-evaluation, where $k$ is the size of a largest minimal
inconsistent subset of $X$.\end{theorem}

The reasoning here is similar to the previous one. For $T$ to violate
rejection-consistency, the set of negations of $T$'s non-members
(drawn from $X$) must be inconsistent. But this set must include
a minimal inconsistent subset, which is of size of most $k$. These
$k$ (or fewer) sentences will each have been deemed true by fewer
than $\frac{1}{k}$ among the $n$ perspectives, which means that
their negations will each have been deemed true by more than $\frac{k-1}{k}$
among the $n$ perspectives (recall that perspectival truth-evaluations
are all classical). So, at least one perspective will have deeemed
all of them true, which once again implies that they cannot be jointly
inconsistent. In analogy to the previous result, if we wanted to achieve
rejection-closedness and not just rejection-consistency, we would
have to push the acceptance quota all the way down to just one perspective,
which would yield subvaluationism.

\subsection{Relaxing unrestricted domain}

Recall that unrestricted domain requires that the overall truth-evaluation
be well-defined for any possible profile of classical truth-evaluations
across the different perspectives. Suppose, by contrast, that there
is some partial agreement among the different perspectives, or that
the extent of their disagreement satisfies certain structural constraints.
It turns out that we may then be able to define overall truth via
the majority function and still achieve a classical truth-evaluation
(at least if we set aside majority ties). There are several domain-restriction
conditions that can do this job (Dietrich and List 2010), but two
particularly interesting ones are the following. \bigskip{}

\noindent\textbf{Value-restricted domain}. The domain of admissible
inputs to the aggregation function consists of every profile $\left\langle T_{1},T_{2},...,T_{n}\right\rangle $
of classical truth-evaluations for which the following holds: for
every minimal inconsistent subset $Y\subseteq X$ there exists a pair
of sentences $\phi,\psi\in Y$ which none of the truth-evaluations
in $\left\langle T_{1},T_{2},...,T_{n}\right\rangle $ deems jointly
true (i.e., every truth-evaluation excludes at least one of $\phi$
and $\psi$ from its set of true sentences). \bigskip{}

\noindent\textbf{Unidimensionally aligned domain}. The domain of
admissible inputs to the aggregation function consists of every profile
$\left\langle T_{1},T_{2},...,T_{n}\right\rangle $ of classical truth-evaluations
for which the following holds. There is a way of aligning the $n$
perspectives on a single axis, from left to right, such that, for
every sentence $\phi$ in $X$, the perspectives deeming $\phi$ true
(i.e., for which $\phi\in T_{i}$) lie either all to the left or all
to the right of those deeming $\phi$ not true (i.e., for which $\phi\in T_{i}$).

\noindent\begin{theorem}Given a value-restricted domain or a unidimensionally
aligned domain, the majority function guarantees a consistent (and
absent majority ties, even classical) overall truth-evaluation, while
also satisfying global unanimity preservation, sentence-wise independence,
and sensitivity.\end{theorem}

We can then define overall truth as truth according to a majority
of perspectives without running into any inconsistencies. Why is this?
First, suppose the profile $\left\langle T_{1},T_{2},...,T_{n}\right\rangle $
is value-restricted. Assume, for a contradiction, the majority truth-evaluation
$T$ is inconsistent. Again, $T$ must include some minimal inconsistent
subset; call it $Y$. But, because the given profile is value-restricted,
$Y$ will contain a pair of sentences $\phi,\psi$ which none of the
truth-evaluations in $\left\langle T_{1},T_{2},...,T_{n}\right\rangle $
deems jointly true. There can then not be majorities for each of $\phi$
and $\psi$, because there would have to be at least one perspective
that jointly ``votes'' for both of them (any two distinct majorities
must overlap). So, $\phi$ and $\psi$ cannot be both contained in
$T$, after all. This completes our proof by contradiction. 

Next, suppose the profile $\left\langle T_{1},T_{2},...,T_{n}\right\rangle $
is unidimensionally aligned. It is then easy to see that none of the
sentences can individually command a majority among the different
perspectives unless it is supported by whichever perspective occupies
the median position (``middlemost'') in the given alignment of the
perspectives from left to right (or unless it is supported by the
two perspectives that lie in the middle, if $n$ is even). It then
follows that the overall truth-evaluation will coincide with the truth-evaluation
of the median perspective or with the intersection of the truth-evaluations
of the median pair of perspectives. The classicality of the individual
perspectives now guarantees the consistency of the resulting majority
truth-evaluation.

The scenario of undimensional alignment might be relevant to the case
of vagueness. Suppose that the perspectival disagreement stems from
different ways of drawing the cut-off line between the satisfaction
and the non-satisfaction of a single vague predicate, such as ``bald''
versus ``not bald'', or ``heap'' versus ``no heap''. It is then
at least conceivable that the effect of this may be a unidimensionally
aligned structure with the perspectives being aligned on a single
axis in an order given by those perspectives' cut-off point for a
relevant vague predicate. Still, unidimensional alignment is a very
demanding condition, and there is no reason to expect that in more
complicated cases, with multiple vague predicates or different sources
of perspectival disagreement, perspectival truth will have this unidimensional
structure.

\section{Extensions}

Before concluding, we would like to comment on, and lift, two technical
assumptions behind our analysis. The first is that the set $N$ of
perspectives is finite. We would like to ask what happens if there
are infinitely many perspectives. The second assumption is that the
relation between perspectival truth and overall truth is captured
by a \emph{function} that admits different possible inputs and correspondingly
produces different possible outputs. The different perspectives' verdicts
are thus allowed to vary from one possible situation to another, with
the overall truth-evaluation varying accordingly. We would like to
ask what happens if there is just a single, fixed profile of truth-evaluations
instead of a domain of different possible ones. In what follows, we
will address these questions in turn.

\subsection{From finitely to infinitely many perspectives}

So, what happens if there are infinitely many perspectives in $N$?
Remarkably, our characterizations of supervaluationism and subvaluationism
(Theorems 1 and 2) continue to work, even though the argument for
them becomes more subtle. The first thing to note is that our first
four conditions on the aggregation function -- unrestricted domain,
consequence-closed truth, global unanimity preservation, and sentence-wise
independence -- still imply that the aggregation function can be
represented in terms of a collection of winning coalitions $\mathcal{W}$
that qualifies as a \emph{filter} on the set of perspectives, namely
a set of subsets of $N$ that satisfies non-triviality, superset-closedness,
and closedness under finite intersections. Crucially, however, it
no longer follows that the intersection of all winning coalitions
must be a winning coalition itself. This is because filters on infinite
sets need not be \emph{principal}. A \emph{principal} filter consists
of all subsets of $N$ that are each supersets of some non-empty subset
$M\subseteq N$. In the infinite case, unlike in the finite one, the
filter need not be ``generated'' by a single set of ``oligarchic''
perspectives. An example of a \emph{non-principal} filter on $N$,
in the infinite case, is the set of all subsets of $N$ whose complement
in $N$ is finite, i.e., 
\[
\mathcal{W}=\left\{ S\subseteq N:\left|N\backslash S\right|<\infty\right\} .
\]
In this example, the intersection of all elements of $\mathcal{W}$
is empty, and thus not in $\mathcal{W}$.\footnote{There can also be non-principal filters where the intersection of
all elements of $\mathcal{W}$ is non-empty but still not contained
in $\mathcal{W}$.} If the aggregation function were defined in terms of this set of
winning coalitions, a sentence would be deemed overall true if and
only if it is true according to all but finitely many perspectives.
The first four conditions of Theorem 1 permit this aggregation function.
But interestingly, our sensitivity condition rules out all non-principal
filters, and, as before, permits only one unique principal filter,
namely the one generated by $M=N$, where $N$ is the only winning
coalition. This is of course supervaluationism. The appendix contains
a full formal proof. Our characterization of subvaluationism generalizes
in a similar way, for reasons of duality. Finally, Theorem 3, which
asserts the non-existence of any aggregation function satisfying all
of our conditions while outputting a classical truth-evaluation, also
continues to hold. 

By contrast, the characterization of dictatorship, namely Theorem
4, which was relevant to the epistemic theory of vagueness, no longer
goes through. The theorem's conditions, which include the requirement
of classical truth, still imply that the aggregation function's collection
of winning coalitions is an ultrafilter on $N$, i.e., a filter on
$N$ that is additionally complete. But the conditions now permit
two kinds of ultrafilters and not just one: principal and free ultrafilters.
A \emph{principal} ultrafilter, as before, is generated by a single
element of $N$, i.e., a ``dictatorial'' perspective, but a \emph{free
}one is a collection of subsets of $N$ with empty intersection (despite
satisyfing closedness under finite intersection). While there are
no free ultrafilters on finite sets, there can be free ultrafilters
on infinite sets, and those would induce an aggregation function satisfying
all of Theorem 4's conditions without being dictatorial.\footnote{As is well known, the relevant possibilities are non-constructive,
i.e., we cannot explicitly construct free ultrafilters, even though
we can deduce their existence, using the axiom of choice.} We leave it for future discussion how such general \emph{ultrafilter
functions} relate to a Williamson-inspired epistemic theory of vagueness
when the number of competing perspectives is infinite.

\subsection{From several admissible profiles to a single one}

We have assumed that the aggregation function admits different possible
profiles of truth-evaluations as input and assigns to each of them
a corresponding truth-evaluation as output. This captures the idea
that the outputs systematically co-vary with the inputs and indeed
supervene on them: no change is possible on the output side without
some change on the input side. Moreover, insofar as we are truth-evaluating
contingent sentences, it is reasonable to think that their truth-values
could vary across different possible situations. 

Let us nonetheless ask what happens if we depart from this ``variable-input''
format and assume instead that there is just a single fixed profile
of truth-evaluations across the different perspectives. This would
amount to switching from the standard ``multi-profile'' approach
in aggregation theory to a ``single-profile'' approach (cf. Roberts
1980). Interestingly, a version of our main characterization theorem
can be derived even in this single-profile setting. 

Consider a fixed profile of classical truth-evaluations, $\left\langle T_{1},T_{2},...,T_{n}\right\rangle $,
and a single corresponding overall truth-evaluation $T$. We call
$T$ \emph{supervaluationist} if $T=\underset{i\in N}{\bigcap}T_{i}$
and, more generally, \emph{oligarchic }if $T=\underset{i\in M}{\bigcap}T_{i}$
for some non-empty set of perspectives $M\subseteq N$. Let us introduce
the following conditions:

\bigskip{}

\noindent\textbf{Consequence-closed truth}. The overall truth-evaluation
$T$ is consequence-closed. 

\bigskip{}

\noindent\textbf{Consistent truth}. The overall truth-evaluation
$T$ is consistent. 

\bigskip{}

\noindent\textbf{Sentence-wise neutrality}. For any $\phi,\psi\in X$,
if $\left[\phi\in T_{i}\Leftrightarrow\psi\in T_{i}\mbox{ for all }i\in N\right]$
then $\left[\phi\in T\Leftrightarrow\psi\in T\right]$.

\bigskip{}

\noindent\textbf{Local sensitivity.} Each perspective $i$ is ``pivotal''
on at least some sentences, i.e., there exists at least one pair of
sentences $\phi,\psi\in X$ on which the overall truth-evaluation
$T$ differs and on which $T_{i}$ is the only perspectival truth-evaluation
that differs.

\bigskip{}

While consequence-closed truth and consistent truth perfectly match
the earlier equally named conditions, sentence-wise neutrality and
local sensitivity can be viewed as single-profile analogues of sentence-wise
independence and sensitivity, respectively. We do not wish to defend
these conditions here, but simply use them to derive an analogue of
our earlier result: 

\begin{theorem}The overall truth-evaluation satisfies consequence-closed
truth, consistent truth, and neutrality if and only if it is oligarchic.
If it further satisfies local sensitivity, it is supervaluationist.\end{theorem}

A dual characterization of subvaluationism could be derived similarly.
This shows that our main findings are, in some sense, robust to a
switch from the multi-profile framework to the single-profile one,
even though we do not advocate this switch.

\section{Concluding remarks}

We have seen that if we impose some \emph{prima facie} plausible conditions
on the relation between perspectival truth and overall truth, then
supervaluationism and subvaluationism are the unique functions for
aggregating the former into the latter. Neither guarantees a classical
truth-evaluation, insofar as they permit truth-value gaps or truth-value
gluts, respectively. If we insist that overall truth should be bivalent,
then the proposed conditions are no longer satisfiable. But if we
relax sensitivity (thereby allowing that not all perspectives matter),
we end up with a dictatorship of one perspective. This last result
arguably captures the structure of the epistemic theory of vagueness. 

Our results have precursors in social choice theory. Our characterization
of supervaluationism is a special case of an axiomatic characterization
of oligarchic aggregation rules in the context of judgment aggregation,
which, in turn, is a generalized variant of a famous characterization
of oligarchic voting rules in social choice theory due to Allan Gibbard
(the result was published as Gibbard 2014 but goes back many decades).
Our characterization of dictatorships, on the other hand, is an application
of the judgment-aggregation variant of Arrow's famous impossibility
theorem (1951/1963) in social choice theory. In its original form,
Arrow's theorem shows that some seemingly reasonable conditions on
the aggregation of preferences imply that one voter must be dictatorial.
Given the ``dictatorial'' structure of the epistemic theory of vagueness,
we have thus identified some novel connections between that theory
and social choice theory.

Finally, our characterizations of supervaluationism and subvaluationism
are robust to the relaxation of two technical assumptions: first,
that the number of competing perspectives is finite, and second, that
we are operating in a multi-profile rather than a single-profile framework
of aggregation. We hope to have provided a novel and robust perspective
on supervaluationism that will inspire future work.

\section*{References}
\begin{lyxlist}{00}
\item [{Arrow,}] K. (1951/1963). \emph{Social Choice and Individual Values}.
New York: Wiley.
\item [{Burnett,}] D. G. (2007). \emph{Trying Leviathan: The Nineteenth-Century
New York Court Case That Put the Whale on Trial and Challenged the
Order of Nature}. Princeton: Prineton University Press.
\item [{Cobreros,}] P., P. Egré, E. Ripley, and R. van Rooij (2012). ``Tolerance
and Mixed Consequence in the S\textquoteright valuationist Setting''.
\emph{Studia Logica} 100(4): 855--877.
\item [{Cobreros,}] P., and L. Tranchini (2014). ``Supervaluationism:
Truth, Value and Degree Functionality.'' \emph{Thought} 3(2): 136--144.
\item [{Dietrich,}] F. (2006). ``Judgment aggregation: (Im)possibility
theorems.'' \emph{Journal of Economic Theory} 126(1): 286--298.
\item [{Dietrich,}] F. (2007). ``A generalised model of judgment aggregation.''
\emph{Social Choice and Welfare} 28(4): 529--565.
\item [{Dietrich,}] F., and C. List (2007a). ``Arrow\textquoteright s
theorem in judgment aggregation.'' \emph{Social Choice and Welfare}
29(1): 19--33.
\item [{Dietrich,}] F., and C. List (2007b). ``Judgment aggregation by
quota rules: Majority voting generalized.'' \emph{Journal of Theoretical
Politics} 19(4): 391--424.
\item [{Dietrich,}] F., and C. List (2008). ``Judgment aggregation without
full rationality.'' \emph{Social Choice and Welfare} 31(1): 15--39.
\item [{Dietrich,}] F., and C. List (2026). \emph{Judgment Aggregation}.
Cambridge: Cambridge University Press.
\item [{Dietrich,}] F., and P. Mongin (2007). ``The Premiss-Based Approach
to Logical Aggregation.'' Unpublished manuscript.
\item [{Dokow,}] E., and R. Holzman (2010a). ``Aggregation of binary evaluations.''
\emph{Journal of Economic Theory} 145(2): 495--511.
\item [{Dokow,}] E., and R. Holzman (2010b). ``Aggregation of binary evaluations
with abstentions.'' \emph{Journal of Economic Theory} 145(2): 544--561.
\item [{Fine,}] K. (1975). ``Vagueness, truth and logic.'' \emph{Synthese}
30(3-4): 265--300.
\item [{Gibbard,}] A. (2014). ``Social choice and the Arrow conditions.''
\emph{Economics and Philosophy} 30(3): 269--284.
\item [{Grinsell,}] T. W. (2019). ``Semantic Indecision.'' In R. Dietz
(ed.), \emph{Vagueness and Rationality in Language Use and Cognition}.
Language, Cognition, and Mind, vol. 5, pp. 135--152. Cham: Springer.
\item [{Hyde,}] D. (1997). ``From Heaps and Gaps to Heaps of Gluts.''
\emph{Mind} 106(424): 641--660.
\item [{Ja\'{s}kowski,}] S. (1969). ``Propositional calculus for contradictory
deductive systems.'' \emph{Studia Logica} 24(1): 143--157.
\item [{Keefe,}] R. (2000). \emph{Theories of Vagueness}. Cambridge: Cambridge
University Press.
\item [{Knight,}] K. (2002). ``Measuring Inconsistency.'' \emph{Journal
of Philosophical Logic }31(1): 77--98.
\item [{List,}] C., and P. Pettit (2002). ``Aggregating sets of judgments:
An impossibility result.'' \emph{Economics and Philosophy} 18(1):
89--110.
\item [{Marcos,}] J., A. P\v{r}enosil, and P. Egré (2026). ``Many-Valued
Logic.'' In E. N. Zalta and U. Nodelman (eds.), \emph{Stanford Encyclopedia
of Philosophy} (Winter 2026 Edition), \textless https://plato.stanford.edu/archives/win2026/entries/logic-manyvalued/\textgreater .
\item [{Pigozzi,}] G. (2006). ``Belief merging and the discursive dilemma:
An argument-based account to paradoxes of judgment aggregation.''
\emph{Synthese} 152(2): 285--298.
\item [{Ripley,}] E. (2013). ``Sorting out the Sorites.'' In K. Tanaka,
F. Berto, E. Mares, and F. Paoli (eds.), \emph{Paraconsistency: Logic
and Applications}, pp. 329--348. Dordrecht: Springer.
\item [{Roberts,}] K. W. S. (1980). ``Social Choice Theory: The Single-Profile
and Multi-Profile Approaches.'' \emph{Review of Economic Studies}
47(2): 441--450.
\item [{Sainsbury,}] M. (2014). ``Fishy business.'' \emph{Analysis} 74(1):
3--5.
\item [{Sorensen,}] R. (2023). ``Vagueness.'' in E. N. Zalta and U. Nodelman
(eds.), \emph{Stanford Encyclopedia of Philosophy} (Winter 2023 Edition),
\linebreak{}
\textless https://plato.stanford.edu/archives/win2023/entries/vagueness/\textgreater .
\item [{Tye,}] M. (1989). \textquotedblleft Supervaluationism and the law
of excluded middle.\textquotedblright{} \emph{Analysis} 49(3): 141--143.
\item [{Varzi,}] A. (1997). ``Inconsistency without contradiction.''
\emph{Notre Dame Journal of Formal Logic} 38(4): 621--639.
\item [{Varzi,}] A. (2007). ``Supervaluationism and its logics.'' \emph{Mind}
116(463): 633--676.
\item [{Williamson,}] T. (1994). \emph{Vagueness}. London: Routledge.
\end{lyxlist}

\appendix

\section{Technical Appendix}

While several of our results are suitably reinterpreted corollaries
of existing results from judgment-aggregation theory (especially the
characterizations of oligarchic and dictatorial aggregation rules),
several others are new, including the results on subvaluationism and
the duality of aggregation functions, the extension of some of our
theorems to infinitely many perspectives and our illustrative result
in the single-profile setting. 

Theorem 1 is a corollary of the main result from Dietrich and List
(2008), which results from adding the sensitivity condition. The sensitivity
condition itself seems to be new in judgment-aggregation theory. Theorem
2 is new. Theorem 3 is a corollary of the main result from Dietrich
and List (2007a), again with the sensitivity condition added. Theorem
4 is a restatement the main result from Dietrich and List (2007a)
and Dokow and Holzman (2010). Theorem 5 is a corollary of a result
from Dietrich (2006), again with the additional sensitivity condition.
Theorem 6 comes from Dietrich and List (2007b). Theorem 7 is new.
Theorem 8 comes from Dietrich and List (2010). Theorem 9 is new.

This appendix provides the new formal proofs, devoting one subsection
to each result. A final subsection will clarify the paper's richness
assumption about the set $X$ of sentences under evaluation, and mention
possible relaxations. The appendix will call the set $X$ the ``agenda'',
in line with the standard terminology in judgment-aggregation theory. 

\subsection{Reducing Theorem 2 to Theorem 1}

We will prove Theorem 2 and (in the next subsection) Theorem 7 by
reducing these results to Theorems 1 and 6, respectively. We denote
the complement of a truth-evaluation by $T\subseteq X$ by $\overline{T}=X\backslash T$,
and the sentence-wide negation of $T$ by $\neg T=\{\neg\phi:\phi\in T\}$.
We define the \emph{dual} of $T$ as $T^{\ast}=\{\lnot p:p\in X\backslash T\}=\neg\overline{T}$.
Note that $T^{\ast\ast}=T$, and that $T^{\ast}=T$ if $T$ is classical.
The set of all classical truth-evaluations is denoted ${\cal J}$.
The \emph{dual} of an aggregation function $F$ with a domain $\mathcal{D}$
is the aggregation function $F^{\ast}$ with the (dual) domain $\mathcal{D}^{\ast}=\{\langle T_{1}^{\ast},\dots,T_{n}^{\ast}\rangle:\langle T_{1},\dots,T_{n}\rangle\in\mathcal{D}\}$
given by:
\[
F^{\ast}(T_{1}^{*},...,T_{n}^{*})=[F(T_{1},...,T_{n})]^{\ast}\text{ for all }\langle T_{1}^{*},...,T_{n}^{*}\rangle\in\mathcal{D}^{*}\text{.}
\]
In $F$ has the unrestricted domain $\mathcal{D}={\cal J}^{n}$, then
the dual domain is still $\mathcal{D}^{\ast}=\mathcal{J}^{n}$, and
$F^{*}$is simply given by
\[
F^{\ast}(T_{1},...,T_{n})=[F(T_{1},...,T_{n})]^{\ast}\text{ for all }\langle T_{1},...,T_{n}\rangle\in\mathcal{J}^{n}\text{.}
\]

The reduction of Theorem 2 to Theorem 1 rests on two lemmas.
\begin{lem}
\label{L1}Supervaluationism and subvaluationism are the duals of
one another.
\end{lem}
\emph{Proof}. Let $F$ be supervaluationism and $G$ subvaluationism.
We must show that $F^{*}=G$ (which implies that $G^{*}=F$ as $F^{**}=F$).
This holds because, for all $\langle T_{1},...,T_{n}\rangle\in\mathcal{J}^{n}$,
\begin{align*}
F^{\ast}(T_{1},...,T_{n}) & =(F(T_{1},...,T_{n}))^{\ast}=(T_{1}\cap\cdots\cap T_{n})^{\ast}=\neg(\overline{T_{1}\cap\cdots\cap T_{n}})\\
 & =\neg(\overline{T_{1}}\cup\cdots\cup\overline{T_{n}})=(\neg\overline{T_{1}})\cup\cdots\cup(\neg\overline{T_{n}})=T_{1}\cup\cdots\cup T_{n},
\end{align*}
where on the second line the first identity holds by de Morgan's law
and the last identity holds because each $T_{i}$ contains exactly
one member of each pair $\phi,\neg\phi\in X$. $\blacksquare$
\begin{lem}
\label{L2}For any aggregation function $F$,
\begin{itemize}
\item $F$ is consequence-closed if and only if its dual $F^{*}$ is rejection-closed,
\item $F$ satisfies unrestricted domain if and only if its dual $F^{*}$does
so,
\item $F$ is independent if and only if its dual $F^{*}$ is so,
\item $F$ is sensitive if and only if its dual \textup{$F^{*}$} is so,
\item $F$ is unanimity-preserving if and only if its dual $F^{*}$ is so.
\end{itemize}
\end{lem}
\emph{Proof}. Consider an aggregation function $F$ on $\mathcal{J}^{n}$.
The last three equivalences are easily shown. The first equivalence
holds because, for any truth-evaluation $T\subseteq X$, $T$ belongs
to the co-domain of $F$ if and only if its dual $T^{*}$ belongs
to the co-domain of $F^{*}$, and $T$ is consequence-closed if and
only if $T^{*}$ is rejection-closed. $\blacksquare$

\medskip{}

Given Lemmas \ref{L1} and \ref{L2}, Theorems 1 and 2 are equivalent
to one another. For instance, if Theorem 1 holds, then Theorem 2 holds
because an aggregation function $F$ satisfies the conditions in Theorem
2 if and only if its dual $F^{*}$ satisfies the conditions in Theorem
1, by Lemma \ref{L2}; equivalently, $F^{*}$ is supervaluationism,
by Theorem 1; equivalently, $F$ is subvaluationism, by Lemma 1.

\subsection{Reducing Theorem 7 to Theorem 6}

The reduction of Theorem 7 to Theorem 6 again rests on two lemmas.
\begin{lem}
\label{L3}For any number $q\in(0,1)$ such that $nq$ is not an integer,
the quota function with quota $q$ and the quota function with quota
$1-q$ are the duals of one another.
\end{lem}
\emph{Proof}. Consider such a $q$, and let $F$ and $G$ be the quota
functions for quota $q$ and $1-q$, respectively. We show that $F^{*}=G$
(which also implies that $G^{*}=F$ since $F^{**}=F$). This holds
because, for all $\langle T_{1},\dots,T_{n}\rangle\in{\cal J}^{n}$
and all $\phi\in X$,
\begin{align*}
\phi\in F^{*}(T_{1},\dots,T_{n}) & \Leftrightarrow\neg\phi\notin F(T_{1},\dots,T_{n})\\
 & \Leftrightarrow|\{i:\neg\phi\in T_{i}\}|<nq\\
 & \Leftrightarrow n-|\{i:\phi\in T_{i}\}|<nq\\
 & \Leftrightarrow|\{i:\phi\in T_{i}\}|>n-nq\\
 & \Leftrightarrow|\{i:\phi\in T_{i}\}|\geq n-nq\text{ (\ensuremath{=n(1-q)})}\\
 & \Leftrightarrow\phi\in G(T_{1},\dots,T_{n}),
\end{align*}
where the penultimate equivalence holds since $|\{i:\phi\in T_{i}\}|\neq n-nq$
as $|\{i:\phi\in T_{i}\}|$ is an integer while $n-nq$ is not an
integer. $\blacksquare$
\begin{lem}
\label{L4}Any aggregation function $F$ guarantees a consistent overall
truth-evaluation if and only if its dual $F^{*}$ guarantees a rejection-consistent
overall truth-evaluation.
\end{lem}
\emph{Proof.} Given $F$, the equivalence holds because, for any truth-evaluation
$T\subseteq X$, $T$ belongs to the co-domain of $F$ if and only
if its dual $T^{*}$ belongs to the co-domain of $F^{*}$, and $T$
is consistent if and only if $T^{*}$ is rejection-consistent. $\blacksquare$\medskip{}

Given Lemmas \ref{L3} and \ref{L4}, Theorems 6 and 7 are equivalent.
For instance, if Theorem 6 holds, then Theorem 7 holds by the following
argument. Let $F$ be a quota function for a quota $q$ in $(0,1)$
such that $q\leq\frac{1}{k}$, where $k$ is defined as in both theorems.
We may assume without loss of generality that $q<\frac{1}{k}$ and
that $nq$ is not an integer, since otherwise we can reduce $q$ slightly
such that both properties hold without affecting the corresponding
quota function. The dual $F^{*}$ is the quota function with quota
$1-q$, by Lemma \ref{L3}. Since $q<\frac{1}{k}$, we have $1-q>\frac{k-1}{k}$.
Thus $F^{*}$ guarantees a consistent overall truth-evaluation, by
Theorem 6. Hence, $F$ guarantees a rejection-consistent overall truth-evaluation,
by Lemma \ref{L4}.

\subsection{Why Theorems 1 and 2 extend to infinitely many perspectives}

Theorems 1 and 2 continue to hold if the set of perspectives $N$
can be infinite. This of course presupposes that we restate the setting
and the definitions of conditions for a possibly infinite set $N$;
the restatement is trivial.\footnote{The set $N$ is now an arbitrary set with $|N|\geq2$. A profile is
now a (possibly infinite) combination $\langle T_{i}\rangle_{i\in N}$
of truth-evaluations, one for each perspective. The definitions of
consequence-closed truth and rejection-closed truth remain unchanged.
In the definition of unrestricted domain, it suffices to replace the
finite profile $\langle T_{1},\dots,T_{n}\rangle$ with $\langle T_{i}\rangle_{i\in N}$;
and analogously for the definitions of sentence-wise independence
and sensitivity. The definition of global unanimity preservation generalizes
as follows: for any profile $\langle T_{i}\rangle_{i\in N}$ in the
domain, if all $T_{i}$ are the same truth-evaluation, then $T$ is
this truth-evaluation.}

How does the proof of Theorem 1 generalize? For reasons like in the
finite case, Theorem 1's conditions (without sensitivity) lead to
a set ${\cal W}$ of winning coalitions that forms a filter over $N$,
i.e., a set of subsets of $N$ that is closed under supersets, closed
under finite intersections, and non-trivial in the sense that $N\in{\cal W}$
and $\emptyset\not\in{\cal W}$. This yields the following result,
a counterpart of a result in judgment aggregation theory (Dietrich
and Mongin 2007):
\begin{lem}
\label{L5}An aggregation function satisfies unrestricted domain,
consequence-closed truth, global unanimity preservation, and sentence-wise
independence if and only if it is a filter function.
\end{lem}
A \emph{filter function} is an aggregation rule $F$ that is given
by a filter ${\cal W}$ over $N$, in the sense that $T=\{\phi\in X:\{i\in N:\phi\in T_{i}\}\in{\cal W}\}$
for all profiles $\langle T_{i}\rangle_{i\in N}$ of classical truth-evaluations.
Crucially, if $N$ is infinite, a filter ${\cal W}$ need not be \emph{principal}:
it need not take the form ${\cal W}=\{S\subseteq N:M\subseteq S\}$
for a non-empty `generating set' $M\subseteq N$.\footnote{The existence of non-principal filters is much more trivial than that
of non-principal ultrafilters. While principal ultrafilters are not
constructible and exist only via the axiom of choice, non-principal
filters can easily be constructed, as the example in the main text
illustrates.} So, in Lemma \ref{L5} we could not replace `filter function' with
`oligarchic function'. An aggregation function $F$ is \emph{oligarchic}
it is given by a non-empty set of `oligarchs' $M\subseteq N$, in
the sense that $T=\cap_{i\in M}T_{i}$ for all profiles $\langle T_{i}\rangle_{i\in N}$
of classical truth-evaluations.

If we now add the sensitivity condition, then only a single filter
function is left, by this (novel) observation:
\begin{lem}
Supervaluationism is the only sensitive filter function.
\end{lem}
\emph{Proof}. Clearly, supervaluationism is a filter function, with
filter ${\cal W}=\{N\}$. Conversely, let $F$ be a sensitive filter
function with an arbitrary filter ${\cal W}$. Assume for a contradiction
that $F$ is not supervaluationism. Then ${\cal W}$ contains a set
of perspectives $S\neq N$. Pick a perspective $i\in N\backslash S$.
As $F$ is sensitive, there exist a $\phi\in X$ and two profiles
$\langle T_{i}\rangle,\langle T_{i}'\rangle$ of classical truth-evaluations
that differ only in $i$'s truth evaluation such that 
\begin{equation}
\phi\notin F(\langle T_{i}\rangle)\text{ and }\phi\in F(\langle T'_{i}\rangle).\label{Eq1}
\end{equation}
Let $S_{\phi}$ ($S_{\phi}'$) be the set of perspectives that deem
$\phi$ true in the first (second) profile. By (\ref{Eq1}), $S_{\phi}\notin{\cal W}$
and $S_{\phi}'\in{\cal W}$. Since the two profiles differ only in
perspective $i$, $S_{\phi}$ and $S_{\phi}'$ differ only in whether
they contain $i$. So, either $S_{\phi}'=S_{\phi}\cup\{\phi\}$, or
$S_{\phi}=S_{\phi}'\cup\{\phi\}$. The latter is, however, impossible,
since it would imply that $S_{\phi}'\subseteq S_{\phi}$, which cannot
be true since $S_{\phi}\notin{\cal W}$ and $S_{\phi}'\in{\cal W}$
and since ${\cal W}$ is superset-closed. Thus ({*}) $S_{\phi}'=S_{\phi}\cup\{\phi\}$.
Since $S_{\phi}'\in{\cal W}$ and $S\in{\cal W}$, we have ({*}{*})
$S_{\phi}'\cap S\in{\cal W}$. By ({*}) and the fact that $i\notin S$,
we have ({*}{*}{*}) $S_{\phi}'\cap S\subseteq S_{\phi}$. By ({*}{*}),
({*}{*}{*}), and the superset-closedness of ${\cal W}$, we have $S_{\phi}\in{\cal W}$,
a contradiction. $\blacksquare$

\medskip{}

Lemmas 1 and 2 jointly imply Theorem 1, in its generalized version
without a finiteness restriction. The (generalized) Theorem 1 implies
the (generalized) Theorem 2 by the earlier duality argument, which
extends immediately to the infinite case.

\subsection{Aggregating truth-evaluations in a single-profile framework}

Most of our results were stated in the standard multi-profile setting,
which considers a domain of possible profiles of truth-evaluations
rather than a single profile. As mentioned however, one could alternatively
work in a single-profile setting. This appendix defines the single-profile
setting more precisely and proves our illustrative result.

The agenda $X$ and the set of perspectives $N$ stay as before ($N$
is kept finite for simplicity). We now consider a fixed profile of
truth-evaluations $\langle T_{1},\dots,T_{n}\rangle$ and a single
corresponding overall truth-evaluation $T$. $T$ is \emph{supervaluationist}
if $T=\cap_{i\in N}T_{i}$, and more generally \emph{oligarchic} if
$T=\cap_{i\in M}T_{i}$ for some non-empty set of ``oligarchs''
$M\subseteq N$. Theorem 9 follows from two lemmas.
\begin{lem}
\label{L7}The overall truth-evaluation satisfies the first three
conditions if and only if it is oligarchic.
\end{lem}
\emph{Proof}. If $T$ is oligarchic, then the three conditions obviously
hold. Conversely, assume the three conditions hold. The coalition
supporting a sentence $\phi\in X$ is denoted $C_{\phi}=\{i\in N:\phi\in T_{i}\}$.
A coalition $C\subseteq N$ is ``winning'' if $T$ contains some
(and hence, by neutrality, \emph{all}) sentences $\phi\in X$ such
that $C_{\phi}=C$. Let ${\cal W}$ be the set of winning coalitions.

\emph{Claim 1:} ${\cal W}$ is closed under finite intersections.

Let $C,C'\in{\cal W}$. Write $C=C_{\phi}$ and $C'=C_{\phi'}$ for
$\phi,\phi'\in T$. As $X$ is closed under conjunction, it contains
a conjunction of $\phi$ and $\phi'$, denoted $\phi\wedge\phi'$.
As $T$ is consequence-closed, $\phi\wedge\phi'\in T$. Thus $C_{\phi\wedge\phi'}\in{\cal W}$.
Meanwhile for each perspective $i$, $\phi\wedge\phi'\in T_{i}\Leftrightarrow\phi,\phi'\in T_{i}$,
by classicality. Thus, $C_{\phi\wedge\phi'}=C_{\phi}\cap C_{\phi'}=C\cap C'$.
Hence, $C\cap C\in{\cal W}$. Q.e.d.

\emph{Claim 2:} ${\cal W}$ is closed under instantiated supersets,
i.e., if $C\in{\cal W}$ and $C\subseteq C'\subseteq N$ where $C'$
is instantiated (i.e., of the form $C_{\psi}$ for some $\psi\in X$),
then $C'\in{\cal W}$.

Let $C$ and $C'$ be as specified. Since $C\in{\cal W}$, there is
a $\phi\in T$ such that $C=C_{\phi}$. Since $C'$ is instantiated,
there is also a $\psi\in X$ such that $C'=C_{\psi}$. Since $X$
is closed under conjunctions and negation, and thus under disjunctions,
it contains a disjunction of of $\phi$ and $\psi$, denoted $\phi\vee\psi$
(whose negation is equivalent to a conjunction of $\neg\phi$ and
$\neg\psi$). Since $\phi$ ($\in T$) entails $\phi\vee\psi$ and
$T$ is consequence-closed, $\psi\in T$. Thus $C'\in{\cal W}$. Q.e.d.

\emph{Claim 3:} $\emptyset\notin{\cal W}$ and $N\in{\cal W}$.

Pick any pair $\phi,\neg\phi\in X$. As $X$ is closed under conjunction,
it contains a conjunction of $\phi$ and $\neg\phi$, hence a contradiction,
denoted $\psi$. By classicality of all $T_{i}$, $C_{\psi}=\emptyset$,
and by consistency of $T$, $\psi\notin T$. Thus $\emptyset\notin{\cal W}$.
By an analogous argument, $X$ contains a tautology $\psi'$, where
$C_{\psi'}=N$ and $\psi'\in T$, so that $N\in{\cal W}$. Q.e.d.

\emph{Claim 4:} The set ${\cal W}^{+}=\{C\subseteq N:C\supseteq C'\text{ for some }C'\in{\cal W}\}$
is a filter.

By construction, ${\cal W}^{+}$ is closed under supersets. It is
non-trivial by Claim 3. To show closure under intersection, consider
$C,C'\in{\cal W}$. Pick $\phi,\phi'\in T$ such that $C_{\phi}\subseteq C$
and $C_{\phi'}\subseteq C'$. As $C_{\phi},C_{\phi'}\in{\cal W}$,
we have $C_{\phi}\cap C_{\phi'}\in{\cal W}$, by Claim 1. Since $C\cap C'\supseteq C_{\phi}\cap C_{\phi'}$,
it follows that $C\cap C'\in{\cal W}^{+}$. Q.e.d.

\emph{Claim 5:} $T$ is oligarchic (completing the proof).

Let $M$ be the coalition generating the filter ${\cal W}^{+}$, so
that ${\cal W}^{+}=\{C\subseteq N:M\subseteq C\}$. $T$ is oligarchic
with set of oligarchs $M$, because for all sentences $\phi\in X$
\begin{align*}
\phi\in T\Leftrightarrow C_{\phi}\in{\cal W}\Leftrightarrow & C_{\phi}\in{\cal W}^{+}\Leftrightarrow M\subseteq C_{\phi}\Leftrightarrow\phi\in\cap_{i\in M}T_{i}.
\end{align*}
Here the second equivalence holds because all coalitions in ${\cal W}^{+}\backslash{\cal W}$
are non-instantiated, since instantiated supersets of winning coalitions
are winning by Claim 2. $\blacksquare$
\begin{lem}
\label{L8}If the overall truth-evaluation is oligarchic and locally
sensitive, then it is supervaluationist.
\end{lem}
\emph{Proof.} This is obvious, since an oligarchic $T$ whose set
of oligarchs excludes a perspective would violate local sensitivity
for this perspective. $\blacksquare$

\subsection{Our assumptions about the agenda}

What exactly have we been assuming about the agenda $X$ and what
relaxations are possible? Readers who are happy to think of $X$ as
the entire language $\mathbf{L}$ can skip this discussion. Following
judgment-aggregation theory, an agenda is a non-empty and (here) finite
set $X\subseteq\mathbf{L}$ that is a disjoint union of pairs $\{\phi,\neg\phi\}$.
(We take double negations to cancel out in a slight abuse of notation,
so that $\neg\neg\phi$ stands for $\phi$.\footnote{Technically, when applying `$\neg$' to agenda sentences, we mean
to switch between members of the agenda pairs $\{\phi,\neg\phi\}$.
This uses `$\neg$' as an agenda-interval version of the negation
operator.}) This paper presupposes two additional features of the agenda: 
\begin{itemize}
\item First, $X$ is \emph{not trivially small}, by including more than
one contingent pair $\{\phi,\neg\phi\}$. The set $X$ thus contains
more than two sentences, besides tautologies and contradictions.
\item Second, $X$ is \emph{closed under conjunctions}: whenever $X$ contains
sentences $\phi,\phi'$, then it also contains a conjunction of $\phi$
and $\phi'$, denoted $\phi\wedge\phi'$ , i.e., a sentence that entails
each of $\phi$ and $\phi'$ and is entailed by $\{\phi,\phi'\}$.
Since agendas are (by definition) closed under negation, this also
implies \emph{closure under disjunctions}: whenever $X$ contains
sentences $\phi,\phi'$, then it also contains a disjunction of $\phi$
and $\phi'$, denoted $\phi\vee\phi'$ , equivalent to $\neg(\neg\phi\wedge\neg\psi)$.
Thus the agenda has the structure of a Boolean algebra.
\end{itemize}
In fact, our results continue to hold under weaker conditions on the
agenda known from judgment-aggregation theory. Theorems 1 to 4 extend
to any agendas that are \emph{path-connected} and \emph{pair-negatable}.\footnote{Path-connectedness requires that any two contingent sentences $\phi,\psi\in X$
can be connected via a path of pairwise conditional entailments: there
exists a sequence of sentences $\phi_{1},\dots,\phi_{k}\in X$ such
that $\phi_{1}=\phi$, $\phi_{k}=\psi$, and each $\phi_{t}$ (with
$t<k$) conditionally entails $\phi_{t+1}$, meaning that $\{\phi_{t}\}\cup Y$
entails $\phi_{t+1}$ for some possibly empty set $Y\subseteq X$
that is consistent with $\phi_{t}$ and consistent with $\neg\phi_{t+1}$.
Pair-negatability requires that at least one minimal inconsistent
set $Y\subseteq X$ becomes consistent by replacing two suitably chosen
members by their negations (e.g., $X$ includes a minimal inconsistent
set of type $\{\phi\wedge\psi,\neg\phi\}$ while $\{\neg(\phi\wedge\psi),\phi\}$
is consistent). For details, see Dietrich and List (2026).} Theorem 5 extends to any agendas that are \emph{not trivially small}
and \emph{atom-closed}.\footnote{A sentence $\alpha\in X$ is an \emph{atom} of $X$ if it entails
exactly one member of each sentence-negation pair $\{\phi,\neg\phi\}\subseteq X$,
and $X$ is \emph{atom-closed} if its set of atoms is exhaustive,
i.e., if $\{\neg\alpha\in X:\alpha$ is an atom of $X\}$ is inconsistent.
Closure under conjunctions guarantees this.} Theorems 6 to 8 extend to all agendas whatsoever. And Theorem 9 extends
to any agendas that are \emph{closed under conjunctions}.
\end{document}